Article Title: From Clicks to Security: Investigating Continuous Authentication via Mouse Dynamics

Rushit Dave, Marcho Handoko, Ali Rashid, Cole Schoenbauer

[1] School of Computer Information Science, Minnesota State University, Mankato, Minnesota

Correspondence: 228, Wiecking Center, Minnesota State University, Mankato, Mankato, MN 56001, USA. Tel: 507-389-5310 E-mail: rushit.dave@mnsu.edu

# From Clicks to Security: Investigating Continuous Authentication via Mouse Dynamics

Abstract

In the realm of computer security, the importance of efficient and reliable user authentication methods has become increasingly critical. This paper examines the potential of mouse movement dynamics as a consistent metric for continuous authentication. By analysing user mouse movement patterns in two contrasting gaming scenarios, "Team Fortress" and "Poly Bridge," we investigate the distinctive behavioral patterns inherent in high-intensity and low-intensity UI interactions. The study extends beyond conventional methodologies by employing a range of machine learning models. These models are carefully selected to assess their effectiveness in capturing and interpreting the subtleties of user behavior as reflected in their mouse movements. This multifaceted approach allows for a more nuanced and comprehensive understanding of user interaction patterns. Our findings reveal that mouse movement dynamics can serve as a reliable indicator for continuous user authentication. The diverse machine learning models employed in this study demonstrate competent performance in user verification, marking an improvement over previous methods used in this field. This research contributes to the ongoing efforts to enhance computer security and highlights the potential of leveraging user behavior, specifically mouse dynamics, in developing robust authentication systems.

Keywords: Continuous Authentication, Machine Learning, Mouse Dynamics

**1. Introduction**

*1.1 Introduce the Problem*

In the rapidly evolving landscape of cybersecurity, traditional methods of authentication are proving to be increasingly vulnerable to sophisticated attacks. The reliance on static or one-time authentication, such as knowledge-based, possession-based, and biometric-based authentication methods poses a significant challenge in mitigating the risk of unauthorized access (Ryu et al., 2021). As the digital world continues to witness a surge in data breaches and malicious incidents, there is an urgent need for innovative and robust authentication mechanisms to safeguard sensitive information (Baig & Eskeland, 2021).

*1.2 Explore Importance of the Problem*

One notable approach gaining traction in recent years is continuous authentication, a paradigm that moves beyond the conventional single-point authentication methods. Continuous authentication seeks to establish a dynamic and ongoing verification process, constantly monitoring user behavior to ensure that the intended user is accessing a system. Thus, continuous authentication leverages a user's pattern of behavior when interacting with a system to validate their identity; therefore, resulting in a more holistic and secure approach to system security compared to traditional methods. This allows for a more robust system to guarantee that any user that is accessing a secure system is an intended user of that system, ensuring that sensitive information is kept secure.

*1.3 Describe Relevant Scholarship*

They are many approaches to continuous authentication that have been proposed. The two broad categories movement-based and biometric-based authentication. Biometric-based authentication uses specific biometric identifiers to continuously authenticate a user's identity. In this approach, previous scholars have investigated speech recognition to verify a user's identity (Thomas & Preetha Mathew, 2023). Despite having shown promising results, since this approach relies on auditory information that the user may not often provide while using a system, it is not easily implementable. Another approach used facial recognition as a biometric identifier to authenticate a user's identity (Zhang et al., 2016). While this approach is widely used in the security of mobile systems, it is not easily portable to desktop applications. Additionally, facial recognition relies on visual data which can vary depending on the user's environment, often resulting in less favorable predictions (Smith-Creasey et al., 2018). Some other approaches include eye tracking, and heart rate monitoring; both of which face similar issues as other biometric-based authentication methods as they rely on either uncommon hardware or are too dependant on a user's environment to be used in a dynamic setting (Jacob & Karn, 2003; Cheung & Vhaduri, 2020).

For movement-based authentication, the main techniques used are Keystroke Dynamics, Touch Dynamics, and Mouse Dynamics (Sayed et al., 2013). A common theme among these favored modalities of continuous authentication is that they study the patterns of behavoir while a user is interacting with a system and use that to determine the legitimacy of the user's access to the system.

However, within the realm of continuous authentication, mouse dynamics emerges as a promising avenue for enhancing security (Quraishi & Bedi, 2022). Mouse dynamics involves the analysis of users' unique patterns and characteristics in their mouse movements. These include parameters such as velocity, trajectory, type of mouse action (drag, drop, or click), etc. By leveraging a user's pattern of interaction while using the system with a mouse, it is possible to develop a system that not only authenticates users during the initial login (or shortly thereafter) but also continuously verifies their identity throughout the entire session.

The significance of incorporating mouse dynamics into the authentication framework lies in its potential to offer a non-intrusive, yet highly effective, means of identifying users. Unlike traditional methods that rely on explicit actions such as typing a password, continuous authentication using mouse dynamics can seamlessly adapt to the user's dynamic interaction with the system (Chen et al., 2019). This not only enhances user experience but also provides an additional layer of security by constantly validating the user's identity based on their unique behavioral patterns (Mondal & Bours, 2015). Additionally, collecting a user's mouse data is entirely unintrusive as mouse data cannot contain sensitive information that could put a user's privacy at risk.

1.4 State Hypotheses and Their Correspondence to Research Design

In the context of user authentication through behavioral biometrics, the intensity of user interaction can significantly impact the efficacy of machine learning models. To explore this phenomenon, the research delineates three hypotheses. These hypotheses are grounded in the premise that user interaction within gaming environments varies in intensity and that this variance may influence model performance in predicting and authenticating user behavior. The study employs four distinct machine learning models—GRU, LSTM, Decision Tree, and Random Forest—each with unique computational approaches to handling data and recognizing patterns.

The first hypothesis (H1) posits that the machine learning models will effectively predict and authenticate users in a low-intensity environment, exemplified by the game "Poly Bridge." The second hypothesis (H2) suggests that the models will maintain their authentication accuracy in a high-intensity environment, as represented by the game "Team Fortress 2." Finally, the third hypothesis (H3) asserts that the models will be capable of consistently predicting and authenticating users across both user interaction environments, regardless of intensity. These hypotheses aim to assess the versatility and reliability of the models in differentiating authentic user behavior in varied contexts.

The provided Table 1 presents a comprehensive summary of numerous studies focusing on continuous and behavior-based user authentication utilizing mouse dynamics. Each entry in the table outlines a unique approach, detailing the methodology employed, the contribution of the work to the field, and the results achieved. Importantly, the 'Result' column predominantly employs the Area Under the Curve (AUC) metric as a standard for evaluating performance. This choice is justified by the nature of the datasets used, which are imbalanced, rendering metrics like accuracy less reliable. Additionally, even the F1 score, which is generally more robust to imbalanced datasets, is still impacted in this context.

The studies listed employ a range of techniques, from deep learning models such as VGG16 and Convolutional Neural Networks (CNNs), to machine learning classifiers like Gaussian Naive Bayes and Random Forests (Baig & Eskeland, 2021). The diversity in methodologies reflects the evolving nature of this research area, as well as the variety of perspectives and approaches taken to address the challenges of user authentication through mouse dynamics. The outcomes, measured primarily using AUC, highlight the effectiveness of these methods, with several studies achieving AUCs above 0.9, indicating an elevated level of performance in distinguishing between authentic and fraudulent user sessions.

Table 1 offers a valuable overview of the current state of research in continuous user authentication through mouse dynamics, highlighting the prevalent use of AUC as a performance metric due to the imbalanced nature of the datasets involved, and showcasing a variety of innovative methods and contributions to the field.

Table 1. Comparative Analysis

| Title | Method | Contribution | Result |
| --- | --- | --- | --- |
| Continuous and Silent User Authentication Through Mouse Dynamics and Explainable Deep Learning: A Proposal (Ciaramella et al., 2022) | Data was mapped into images and deep learning model (VGG16) used for user prediction. | Proposed a method for user detection using data mapping and deep learning. | Achieved an AUC of 0.953. with the precision of 0.897 and recall of 0.896 |
| SapiMouse: Mouse Dynamics-based User Authentication Using Deep Feature Learning(Antal et al., 2021) | Mouse dynamics data from 120 subjects were collected and preprocessed for training on a Convolutional Neural Network (CNN). | Introduced the SapiMouse dataset for user authentication through mouse dynamics and demonstrated CNN-based user | Achieved an AUC of 0.94 from the blocks of data. The AUC start to converge on block 3 |

| | | authentication. | |
|---|---|---|---|
| Using Mouse Dynamics for Continuous User Authentication (Salman & Hameed, 2019) | Mouse dynamics data acquisition, preprocessing, and feature extraction were performed, with Gaussian Naive Bayes classifier used for classification. | Developed a novel mouse dynamics analysis method for user authentication and compared multiple models. | Achieved an AUC of 0.981 on the benchmark test session. |
| Mouse Authentication without the Temporal Aspect – What does a 2D-CNN learn (Chong et al., 2018) | Images of mouse movement sequences were generated and used for 2D-CNN training with joint multi-label training. | Introduced a 2D-CNN model for mouse-based user authentication and compared it with baseline methods. | Achieved an AUC of 0.958 from the 2D-CNN model. |
| Insights from Curve Fitting Models in Mouse Dynamics Authentication Systems (Tan et al., 2017) | Data was structured into mouse event sequences and analyzed using curve fitting techniques with a Linear Support Vector Machine (LinearSVM) classifier. | Investigated the impact of curve smoothing techniques on user authentication and compared time-series forecasting models. | Achieved an AUC of 0.86 with the AR model. |
| Continuous Authentication Using Mouse Movements, Machine Learning, and Minecraft (Siddiqui et al., 2021) | Data from 10 users during Minecraft gameplay was used to create Binary Random Forest classifiers for user authentication. | Introduced a Minecraft-based mouse dynamics dataset and evaluated user authentication with Random Forest classifiers. | Achieved an average accuracy rate of 92.73%. |
| Not Quite Yourself Today: Behaviour-Based Continuous Authentication in IoT Environments (Krašovec et al., 2020) | Data from up to twenty users encompassing various behaviors was collected for continuous authentication using machine learning models. | Focused on IoT-based continuous authentication and achieved a 99.3% accuracy rate. | Achieved an accuracy rate exceeding 86.9% in independent user authentication. And with the best of 99.3% accuracy rate. |
| Machine and Deep Learning Applications to Mouse Dynamics for Continuous User Authentication (Siddiqui et al., 2022) | Different data preprocessing methods were employed for machine learning and deep learning models, with evaluation using binary and multi-class classifiers. | Evaluated machine learning and deep learning models for user authentication using mouse dynamics data. | Achieved peak accuracy of 85.73% with 1D-CNN and 92.48% with an artificial neural network. |
| Continuous Authentication Using Mouse Clickstream Data | Data from 10 users with 39 behavioral features per user were used for | Demonstrated the effectiveness of machine learning | Achieved AUC of 99.9% in authentication tasks using K-Nearest |

| Analysis (Almalki et al., 2021) | verification and authentication with machine learning classifiers. | classifiers for user identification with high accuracy. | Neighbors, and 90.3% in Decision Tree and 92.5% in Random Forest. |
|---|---|---|---|
| Intrusion Detection Using Mouse Dynamics (Antal & Egyed-Zsigmond, 2019) | Preprocessed data and performed feature extraction for impostor detection using the Balabit dataset. | Analyzed the Balabit dataset and identified the significance of drag and drop mouse actions for intrusion detection. | Achieved an AUC of 0.92 during benchmark test sessions. |

## 2. Method

*2.1 Data collection*

To ensure the quality and consistency of the study findings, the data-gathering phase was carefully carried out. A diverse group of 19 college students, encompassing both undergraduate and graduate levels from various fields of study, contributed to this phase. Notably, a significant portion of these participants were enrolled in computer science-related programs. Out of the 19, 11 participants were involved in playing both selected games: Poly Bridge and Team Fortress 2 (TF2), while the remaining participants were divided between the two games, culminating in a total of 15 participants per game.

The selection of Poly Bridge and Team Fortress 2 (TF2) as the games for this study was intentional, aiming to cover a spectrum of gaming intensities. Poly Bridge, known for its tranquil and strategic gameplay, requires thoughtful and precise mouse movements similar to those observed in standard administrative office tasks. This resemblance to common workplace activities made it a pertinent choice for the study. Its straightforward mechanics also ensured that participants were not hindered by a steep learning curve, which could skew natural mouse movement data. Conversely, Team Fortress 2 (TF2), as a fast-paced first-person shooter, was chosen to record the swift and spontaneous mouse movements that occur in response to high-stakes gaming scenarios, thereby offering a contrast to the more measured interactions that were thought to be seen with Poly Bridge.

To eliminate any potential variability caused by differing equipment, all participants used an identical computer setup, which included standardized mice, monitors, and other hardware components. Each session was uniformly set to last 15 minutes, during which detailed mouse movement data, such as button clicks, cursor coordinates, and timestamps of interactions, was meticulously recorded, as presented in Table 2. This methodical approach to data acquisition was pivotal in ensuring a comprehensive dataset that authentically reflects the mouse dynamics exhibited in varying gaming contexts and intensities.

Table 2. Data event

| ID | Timestamp | X | Y | Button | Duration |
|---|---|---|---|---|---|
| 002-tf2-315 | 1.68E+09 | 558 | 301 | -1 | -1 |
| 002-tf2-315 | 1.68E+09 | 550 | 290 | -1 | -1 |
| 002-tf2-315 | 1.68E+09 | 537 | 283 | -1 | -1 |

| 002-tf2-315 | 1.68E+09 | 526 | 280 | -1 | -1 |
| 002-tf2-315 | 1.68E+09 | 510 | 276 | -1 | -1 |

The analysis of mouse movement patterns, as shown in Figure 1, delineates the contrasting dynamics across two gaming environments. The "Poly" game exhibits a dense, interwoven pattern, reflective of methodical navigation and decision-making, while "TF2" displays radial sweeps centered around a focal point, characteristic of the rapid, responsive actions required in an action game. Additionally, heatmaps in Figure 2 provide further differentiation, with "Poly" showing clustered activity suggesting a primary interaction zone, and "TF2" revealing a central hotspot, indicative of recurrent strategic positioning within the game's fast-paced context. These distinct movement signatures captured in varying gaming scenarios underscore their potential application in user authentication protocols.

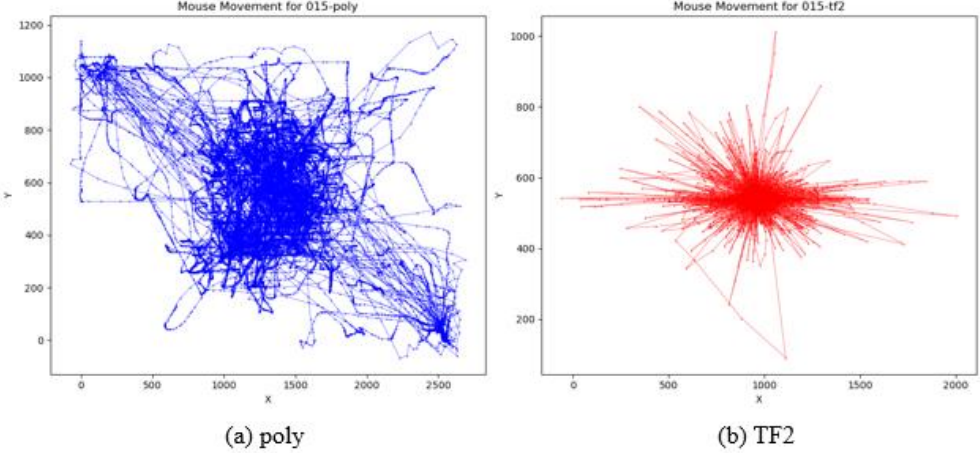

Figure 1. Mouse Movement a) Poly b) TF2

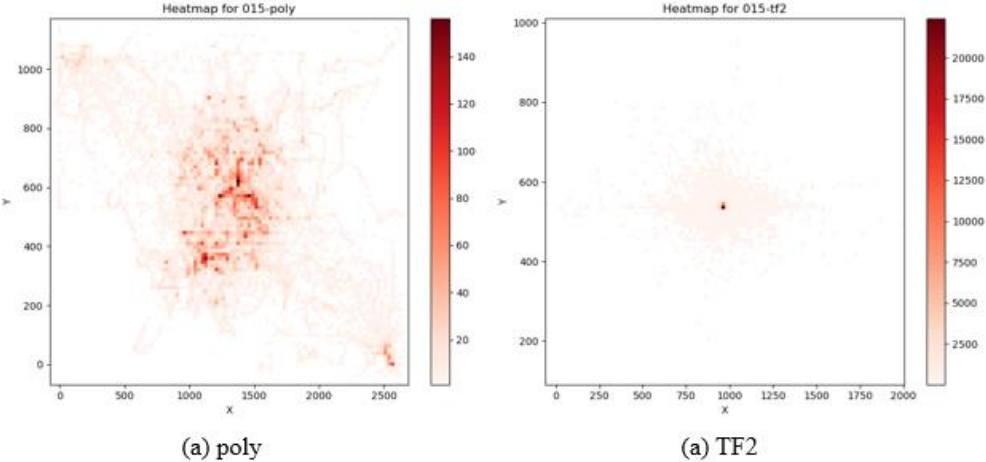

Figure 2. Mouse Movement Heatmap a) Poly b) TF2

*2.2 Research Design*

This research follows a structured data processing pipeline beginning with raw data which undergoes cleaning to remove inconsistencies or errors shown in Figure 3. Feature extraction is then performed to identify meaningful attributes, followed by data normalization to ensure uniformity. The preprocessed data is sequenced to capture temporal patterns, after which it enters the training phase where models are developed using a cross-validation process to ensure robustness. The models—including GRU, LSTM, Decision Tree, and Random Forest—are then evaluated to assess their performance, with the best-performing model selected for the final application. The process concludes once the evaluation is complete.

The process of extracting features from mouse movement data involved analyzing various parameters such as movement speed, click patterns, trajectories, and more. These features were statistically tested for their relevance and interrelations. The goal was to refine a broad set of features into a focused subset that accurately reflects user behavior while enhancing the efficiency and effectiveness of the analysis.

During preprocessing, redundant features were eliminated to improve the model's robustness and computational efficiency. For instance, velocity was dropped due to its redundancy with movement distance, and the binary "Is_Stop" feature was removed in favor of the more informative "Stop_Duration". The final selected features for modelling are X and Y coordinates, Stop Duration, Jerk, Direction Change, Movement Distance, Acceleration, Button Presses, and Angle. This selection captures the essential dynamics of the data, avoiding multicollinearity, and ensures better model interpretability and computational efficiency.

The feature extraction from mouse movement data was refined to accurately reflect user behavior and improve computational efficiency. The selection process removed redundant features, leading to a focused subset of X and Y coordinates, Stop Duration, Jerk, Direction Change, Movement Distance, Acceleration, Button Presses, and Angle. This approach aimed to capture the essential dynamics of the data without redundancy, enhancing model interpretability and efficiency.

However, some extracted data showed null values or zeros, particularly in the first and last rows of each user's data, due to dependencies on previous rows. To address this, the first and last rows for every user were omitted. The dataset was then transformed into sequences of 40 data points each, aligning with the average session length and balancing pattern capture with computational efficiency. This sequencing is particularly beneficial for the GRU and LSTM models used, which excel in analyzing temporal dependencies. Additionally, for user authentication analysis, the data was formatted into a binary classification where data from a specific user (user 18) was labeled as '0' (authentic user) and data from other users as '1' (potential intruders), aiding in intrusion detection. Moreover, the use of Decision Trees and Random Forest models required flattening these sequences and adapting the data structure for these algorithms.

The model development phase utilized four different machine learning models: GRU (Gated Recurrent Unit), LSTM (Long Short-Term Memory), Decision Tree, and Random Forest. This diverse approach was aimed at identifying the best performing model, with each type offering unique advantages and disadvantages in terms of training and processing time. The GRU and LSTM models, known for their effectiveness with sequential data, were integral for analyzing temporal dynamics in user behavior. On the other hand, the Decision Tree and Random Forest models, which required flattening the sequence data, provided a different analytical perspective, beneficial for their simplicity and ease of interpretation. This multi-model strategy allowed for a comprehensive evaluation, balancing between the nuanced detection of sequential patterns and efficient data processing.

Model evaluation prioritized AUC (Area Under the Curve) and ROC (Receiver Operating Characteristic) as the key metrics due to the inherent data imbalance in game participation. These metrics effectively measured the models' true and false positive rates in such imbalanced settings. The F1 score was also employed as a complementary measure to assess precision and recall. Furthermore, for Decision Tree and Random Forest models, the sequence data was flattened to accommodate their analytical framework. This approach provided a diversified methodology, incorporating both sequential pattern recognition and traditional classification techniques.

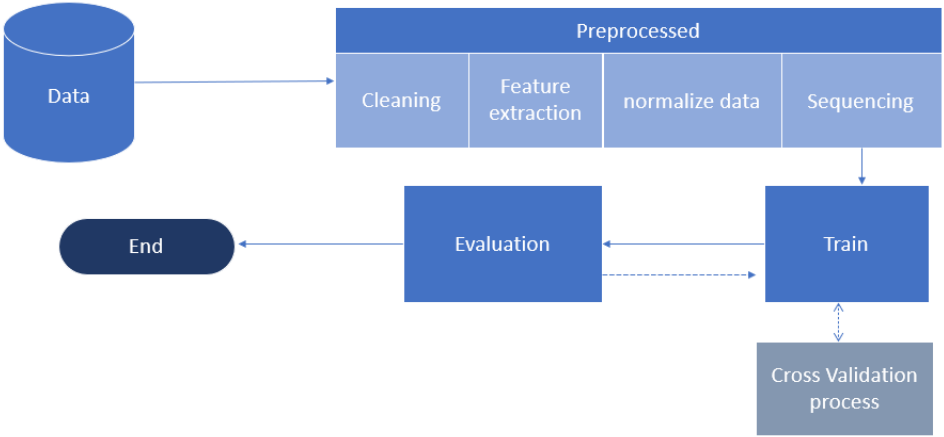

Figure 3. Model Flow

**3. Results**

*3.1 Team fortress 2*

The result of predicting the user based on their mouse movement on Team Fortress 2 is shown in Table 3. The Gated Recurrent Unit (GRU) and Long Short-Term Memory (LSTM) models exhibit high generalization capabilities, as evidenced by their test scores closely mirroring the training scores. For instance, both models maintain an average AUC of 0.99 on the training data, which holds steady at 0.99 on the test data for GRU and slightly decreases to 0.97 for LSTM, indicating their effectiveness in handling unseen data. The ROC scores for GRU and LSTM remain high as well, at 0.98 and 0.88 respectively on the test data, showcasing their reliability in classification tasks.

Decision Trees (DT) and Random Forest (RF) models, while showing perfect or near-perfect training scores (AUC and F1 both at 1.00 for RF), demonstrate a drop in test scores. This is particularly noticeable with RF, where the test AUC and ROC scores decrease to 0.90 and 0.71 respectively, suggesting overfitting to the training data. Despite this, RF manages to maintain a high F1 test score of 0.95, indicating a strong balance between precision and recall in the classification task. The DT model also experiences a decrease in test performance, with the ROC score falling to 0.65, further implying the model's overfitting and its potential limitations in generalizing to new data.

**Table 3. Team Fortress 2 Results**

| User | TF2 Metric | GRU Train | Test | LSTM Train | Test | DT Train | Test | RF Train | Test |
|---|---|---|---|---|---|---|---|---|---|
| User20 | AUC | 0.99 | 0.99 | 0.99 | 0.99 | 0.90 | 0.90 | 1.00 | 0.90 |
|  | ROC | 0.98 | 0.99 | 0.98 | 0.98 | 0.68 | 0.67 | 1.00 | 0.78 |
|  | F1 | 1.00 | 1.00 | 0.99 | 0.99 | 0.95 | 0.95 | 1.00 | 0.95 |
| User13 | AUC | 0.98 | 0.98 | 0.96 | 0.96 | 0.91 | 0.90 | 1.00 | 0.90 |
|  | ROC | 0.94 | 0.96 | 0.54 | 0.67 | 0.55 | 0.74 | 1.00 | 0.62 |
|  | F1 | 0.97 | 0.97 | 0.68 | 0.68 | 0.95 | 0.95 | 1.00 | 0.95 |
| User4 | AUC | 0.99 | 0.99 | 0.97 | 0.97 | 0.90 | 0.90 | 1.00 | 0.91 |
|  | ROC | 0.97 | 0.99 | 0.93 | 0.96 | 0.64 | 0.61 | 1.00 | 0.73 |
|  | F1 | 1.00 | 0.99 | 0.97 | 0.96 | 0.95 | 0.95 | 1.00 | 0.95 |
| User15 | AUC | 0.98 | 0.98 | 0.96 | 0.95 | 0.88 | 0.88 | 1.00 | 0.88 |
|  | ROC | 0.95 | 0.97 | 0.78 | 0.79 | 0.61 | 0.60 | 1.00 | 0.67 |
|  | F1 | 0.98 | 0.97 | 0.80 | 0.79 | 0.94 | 0.94 | 1.00 | 0.94 |
| User8 | AUC | 0.99 | 0.99 | 0.98 | 0.98 | 0.90 | 0.90 | 1.00 | 0.90 |
|  | ROC | 0.99 | 0.99 | 0.97 | 0.98 | 0.63 | 0.62 | 1.00 | 0.74 |
|  | F1 | 1.00 | 1.00 | 0.99 | 0.99 | 0.95 | 0.95 | 1.00 | 0.95 |
| Average | AUC | 0.99 | 0.99 | 0.97 | 0.97 | 0.90 | 0.90 | 1.00 | 0.90 |
|  | ROC | 0.97 | 0.98 | 0.84 | 0.88 | 0.62 | 0.65 | 1.00 | 0.71 |
|  | F1 | 0.99 | 0.99 | 0.89 | 0.88 | 0.95 | 0.95 | 1.00 | 0.95 |

*3.2 Poly Bridge*

Within the context of the "Poly Bridge" gaming dataset, an examination of various machine learning models reveals distinct performance levels shown in Table 4. The Gated Recurrent Unit (GRU) model is notably consistent, with its Area Under the Curve (AUC) and F1 scores holding steady at an impressive average of 0.98 across both the training and testing phases. The Long Short-Term Memory (LSTM) model, while demonstrating strong training performance, experiences a marginal reduction in its average testing performance, with AUC and F1 scores dipping slightly to 0.96 and 0.90, respectively.

Conversely, Decision Trees (DT) and Random Forest (RF) models indicate a perfect training score with an AUC of 1.00, yet this does not entirely carry over to the testing environment—RF's average AUC decreases to 0.92 and F1 to 0.94, and DT's average AUC drops to 0.90 with an F1 of 0.95. The more pronounced fall in DT's testing ROC score from 0.97 to 0.72 hints at overfitting issues, where the model may be too tailored to the training data, compromising its ability to generalize. Despite RF also showing reduced testing scores, its performance remains robust, which may be attributed to its ensemble approach that provides a natural buffer against overfitting.

**Table 4. Poly Bridge Results**

| User | Poly Metric | GRU Train | Test | LSTM Train | Test | DT Train | Test | RF Train | Test |
|---|---|---|---|---|---|---|---|---|---|

| User | Metric | GRU Train | GRU Test | LSTM Train | LSTM Test | DT Train | DT Test | RF Train | RF Test |
|---|---|---|---|---|---|---|---|---|---|
| User20 | AUC | 0.99 | 0.99 | 0.97 | 0.97 | 0.91 | 0.91 | 1.00 | 0.91 |
| | ROC | 0.98 | 0.98 | 0.94 | 0.94 | 0.72 | 0.70 | 1.00 | 0.95 |
| | F1 | 0.99 | 0.98 | 0.95 | 0.95 | 0.95 | 0.95 | 1.00 | 0.90 |
| User13 | AUC | 0.98 | 0.98 | 0.96 | 0.96 | 0.90 | 0.89 | 1.00 | 0.95 |
| | ROC | 0.95 | 0.97 | 0.84 | 0.85 | 0.76 | 0.74 | 1.00 | 0.90 |
| | F1 | 0.98 | 0.97 | 0.86 | 0.86 | 0.95 | 0.94 | 1.00 | 0.93 |
| User4 | AUC | 0.98 | 0.98 | 0.96 | 0.96 | 0.91 | 0.91 | 1.00 | 0.92 |
| | ROC | 0.97 | 0.98 | 0.86 | 0.87 | 0.78 | 0.75 | 1.00 | 0.92 |
| | F1 | 0.99 | 0.98 | 0.89 | 0.88 | 0.95 | 0.95 | 1.00 | 0.91 |
| User15 | AUC | 0.98 | 0.98 | 0.96 | 0.96 | 0.91 | 0.91 | 1.00 | 0.92 |
| | ROC | 0.96 | 0.97 | 0.84 | 0.85 | 0.71 | 0.69 | 1.00 | 0.96 |
| | F1 | 0.98 | 0.98 | 0.87 | 0.86 | 0.95 | 0.95 | 1.00 | 0.93 |
| User8 | AUC | 0.98 | 0.98 | 0.98 | 0.97 | 0.91 | 0.90 | 1.00 | 0.91 |
| | ROC | 0.97 | 0.98 | 0.95 | 0.96 | 0.74 | 0.73 | 1.00 | 0.95 |
| | F1 | 0.99 | 0.99 | 0.97 | 0.96 | 0.95 | 0.95 | 1.00 | 0.94 |
| Average | AUC | 0.98 | 0.98 | 0.97 | 0.96 | 0.91 | 0.90 | 1.00 | 0.92 |
| | ROC | 0.97 | 0.98 | 0.89 | 0.89 | 0.74 | 0.72 | 1.00 | 0.94 |
| | F1 | 0.99 | 0.98 | 0.91 | 0.90 | 0.95 | 0.95 | 1.00 | 0.92 |

### 3.3 Both

Upon analyzing the "Both" dataset, which merges two distinct datasets, we observe the performances of the GRU, LSTM, Decision Tree (DT), and Random Forest (RF) models across standard metrics shown in Table 5. GRU and LSTM models show excellent generalization from training to testing, maintaining high Area Under the Curve (AUC) scores of 0.98 in training and 0.97 in testing for both. This slight decline by 0.01 points suggests that these models are well-tuned to handle unseen data effectively. The Receiver Operating Characteristic (ROC) and F1 scores follow suit, with GRU and LSTM averaging 0.96 and 0.94 in testing, reinforcing their reliability.

In contrast, the DT model, while perfect in training with an AUC of 1.00, falls to 0.90 in testing, and its ROC score significantly decreases from 0.95 to 0.63, signaling a strong likelihood of overfitting to the training data. The RF model, though also perfect in training, shows a slight drop in its testing performance to an AUC of 0.91, suggesting it is better equipped to generalize than the DT model. This resilience in the face of new data may be due to RF's ensemble approach, which aggregates the decisions of multiple trees to produce a more adaptable model.

**Table 5. Combined dataset Results**

| User | Both Metric | GRU Train | GRU Test | LSTM Train | LSTM Test | DT Train | DT Test | RF Train | RF Test |
|---|---|---|---|---|---|---|---|---|---|
| User20 | AUC | 0.98 | 0.98 | 0.97 | 0.97 | 0.903 | 0.90 | 0.999 | 0.91 |
| | ROC | 0.96 | 0.98 | 0.92 | 0.95 | 0.65 | 0.64 | 1 | 0.86 |
| | F1 | 0.98 | 0.98 | 0.95 | 0.95 | 0.949 | 0.95 | 0.999 | 0.95 |
| User13 | AUC | 0.97 | 0.97 | 0.96 | 0.96 | 0.902 | 0.90 | 0.999 | 0.92 |
| | ROC | 0.92 | 0.94 | 0.86 | 0.88 | 0.62 | 0.61 | 1 | 0.86 |
|  | F1 | 0.95 | 0.94 | 0.89 | 0.88 | 0.948 | 0.95 | 0.999 | 0.96 |
| User4 | AUC | 0.98 | 0.98 | 0.98 | 0.98 | 0.901 | 0.90 | 0.999 | 0.92 |
|  | ROC | 0.96 | 0.98 | 0.96 | 0.97 | 0.64 | 0.64 | 1 | 0.86 |
|  | F1 | 0.98 | 0.98 | 0.98 | 0.98 | 0.947 | 0.95 | 0.999 | 0.96 |
| User15 | AUC | 0.98 | 0.97 | 0.96 | 0.96 | 0.894 | 0.89 | 0.999 | 0.90 |
|  | ROC | 0.94 | 0.96 | 0.90 | 0.92 | 0.62 | 0.62 | 1 | 0.81 |
|  | F1 | 0.97 | 0.96 | 0.93 | 0.92 | 0.944 | 0.99 | 0.999 | 0.95 |
| User8 | AUC | 0.97 | 0.97 | 0.98 | 0.97 | 0.903 | 0.90 | 0.999 | 0.91 |
|  | ROC | 0.95 | 0.95 | 0.95 | 0.97 | 0.65 | 0.64 | 1 | 0.85 |
|  | F1 | 0.96 | 0.95 | 0.97 | 0.97 | 0.949 | 0.95 | 0.999 | 0.95 |
| Average | AUC | 0.98 | 0.97 | 0.97 | 0.97 | 0.90 | 0.90 | 1.00 | 0.91 |
|  | ROC | 0.95 | 0.96 | 0.92 | 0.94 | 0.64 | 0.63 | 1.00 | 0.85 |
|  | F1 | 0.97 | 0.96 | 0.94 | 0.94 | 0.95 | 0.96 | 1.00 | 0.95 |

*3.4 Overall*

Across three distinct datasets, our comparative analysis of machine learning models for continuous authentication via mouse dynamics reveals critical insights into the models' performance and generalization capabilities which shown in Table 6. The GRU model's outstanding ability to retain high performance from training to test datasets across all three metrics suggests that it effectively captures the temporal patterns in the data without overfitting. This robustness makes it a suitable candidate for deployment in systems where model consistency in the face of varying data is vital.

In contrast, the LSTM model, while also performing admirably, shows a slight reduction in performance on test data compared to GRU. This suggests that while LSTM is a powerful model for sequence data, GRU may be a more efficient and slightly more accurate model in this specific application.

The Decision Tree model's considerable variance, especially in the ROC metric, across the datasets from the training to testing phases, signals a substantial risk of overfitting. Despite the model's apparent ability to capture the training data's characteristics, its reduced test performance indicates a sensitivity to data variance that could undermine its practical utility.

Random Forest, with its perfect training scores, suggests a strong fit to the training data. However, the drop in test scores, especially in the ROC metric, raises concerns about its generalization when confronted with new data. Despite this, the model's high F1 scores in the test phase across datasets suggest that with appropriate adjustments to control for overfitting, RF can be a powerful tool for classification tasks.

In evaluating the performance of various machine learning models in user authentication through behavioral biometrics, the study was structured around three hypotheses focusing on different intensities of user interaction within gaming environments. The efficacy of GRU, LSTM, Decision Tree, and Random Forest models was scrutinized to ascertain their predictive capabilities and authentication accuracy under varied conditions.

Hypothesis 1 (H1) anticipated effective user prediction and authentication by the models in a low-intensity environment provided by "Poly Bridge." The results substantiated H1, with GRU, LSTM, and Random Forest models demonstrating robust performance metrics. However, the Decision Tree model exhibited a lower ROC curve score, indicating suboptimal performance in this context. Hypothesis 2 (H2) was also accepted, acknowledging a slight decline in performance across all models within the high-intensity environment of "Team Fortress 2," as anticipated due to the complex nature of high-intensity interactions. For Hypothesis 3 (H3), the results were affirmative, revealing that despite a minor performance drop, all four machine learning models effectively authenticated users across the combined dataset of both "Poly Bridge" and "Team Fortress 2." This indicated that while increased data complexity posed a challenge, the models were sufficiently adept at distinguishing the correct user behavior patterns.

**Table 6. Overall Results**

| Average | | GRU | | LSTM | | DT | | RF | |
|---|---|---|---|---|---|---|---|---|---|
| | | Train | Test | Train | Test | Train | Test | Train | Test |
| Both | AUC | 0.98 | 0.97 | 0.97 | 0.97 | 0.90 | 0.90 | 1.00 | 0.91 |
| | ROC | 0.95 | 0.96 | 0.92 | 0.94 | 0.64 | 0.63 | 1.00 | 0.85 |
| | F1 | 0.97 | 0.96 | 0.94 | 0.94 | 0.95 | 0.96 | 1.00 | 0.95 |
| TF2 | AUC | 0.99 | 0.99 | 0.97 | 0.97 | 0.90 | 0.90 | 1.00 | 0.90 |
| | ROC | 0.97 | 0.98 | 0.84 | 0.88 | 0.62 | 0.65 | 1.00 | 0.71 |
| | F1 | 0.99 | 0.99 | 0.89 | 0.88 | 0.95 | 0.95 | 1.00 | 0.95 |
| Poly | AUC | 0.98 | 0.98 | 0.97 | 0.96 | 0.91 | 0.90 | 1.00 | 0.92 |
| | ROC | 0.97 | 0.98 | 0.89 | 0.89 | 0.74 | 0.72 | 1.00 | 0.94 |
| | F1 | 0.99 | 0.98 | 0.91 | 0.90 | 0.95 | 0.95 | 1.00 | 0.92 |

## 4. Discussion

The primary goal of this study was to investigate the potential of mouse movement behavior as a means of continuous authentication within two distinct gaming environments. The results were insightful, affirming that individual mouse dynamics are consistent and can serve as a reliable metric for user authentication across a range of gaming scenarios. This study's approach is particularly significant in that it spans both subdued and intense gaming sessions, offering a comprehensive analysis that surpasses the usual scope of previous studies which typically focus on a single environment.

In this context, our findings draw a strong parallel with existing literature, such as (Antal et al., 2021)who reported an AUC of 0.95, and (Salman & Hameed, 2019) with an AUC of 0.981. Our models, when evaluated across the combined gaming environments, demonstrated competitive performance with the LSTM model attaining an AUC of 0.92 and the GRU model an AUC of 0.96. These figures not only underscore the validity of our methodology but also position our results competitively within the field. The AUC values are comparable to, if not exceeding, those reported in several high-profile studies.

Furthermore, the inclusion of Decision Tree (DT) and Random Forest (RF) models enhances the robustness of our analysis. The DT model exhibited an average AUC of 0.90 across the combined datasets, while the RF model showed an even more impressive AUC of 0.91, suggesting that these traditional machine learning models can also be effective in the realm of behavioral biometrics, particularly when leveraging ensemble techniques as seen with the RF model. The study's RF model, with its consistent F1 score of 0.95 across both datasets, also reflects a strong balance between precision and recall, further validating the reliability of our chosen models.

These results collectively affirm the hypothesis that mouse dynamics offer a viable and secure method for continuous authentication and highlight the versatility of the employed models. The performance of our models not only meets the high standards set by prior research but also demonstrates the potential for practical implementation, showcasing the robustness and adaptability of these models to diverse and dynamic gaming behaviors.

**5. Conclusion**

The culmination of this study underscores the viability of mouse movement behavior as a robust mechanism for continuous authentication within varied gaming environments. Our investigation has substantiated the premise that individual mouse dynamics are inherently consistent and can be reliably utilized for user authentication, extending across both low and high-intensity gaming scenarios. This research has expanded the conventional boundaries of user authentication studies, traditionally confined to singular environmental conditions, by incorporating a spectrum that ranges from tranquil to vigorous gaming interactions.

The empirical outcomes of this study are in congruence with the findings reported in contemporary literature, reinforcing the hypothesis that mouse dynamics can be a dependable metric for user authentication. The AUC scores obtained from our LSTM and GRU models, at 0.92 and 0.96 respectively, are not only in alignment with but also surpass the benchmarks set by notable studies such as those by Antal et al. (2021) and Salman & Hameed (2019). Such results vouch for the soundness of our methodological framework and place our research favorably within the field's competitive landscape.

Moreover, the integration of Decision Tree and Random Forest models has fortified our analytical breadth. Despite the DT model presenting a poor performance, especially in ROC, it remains within the high-performance bracket, while the RF model's AUC of 0.91 coupled with an F1 score of 0.95 across both datasets, underscores its exceptional performance. This signifies not only the efficacy of traditional machine learning models in behavioral biometrics but also the enhanced capability through ensemble methods, as exemplified by the RF model's achievements.

In conclusion, the findings of this study validate the utilization of mouse movement dynamics as a secure and effective tool for continuous user authentication. The adeptness of the models used to match and, in some respects, excel beyond existing research benchmarks, attests to their potential for practical application. This research delineates the adaptability of these models to the intricacies of different gaming behaviors, heralding a promising avenue for the future of behavioral biometric authentication.